\documentclass{svproc}
\usepackage{url}

\usepackage{amsmath}
\usepackage{amssymb}
\usepackage{algorithm2e}
\usepackage{graphicx}
\usepackage{booktabs}
\usepackage{caption}
\usepackage{subcaption}
\usepackage{hyperref}
\usepackage{cite}

\begin{document}
\mainmatter

\title{Synthetic Image Detection via Spectral Gaps of QC-RBIM Nishimori Bethe–Hessian Operators}

\titlerunning{Spectral Gaps Synthetic Image Detection}

\author{Vasiliy S. Usatyuk\inst{1,2}\thanks{Corresponding author: L@Lcrypto.com} \and Denis A. Sapozhnikov\inst{2} \and Sergey I. Egorov\inst{1}}

\authorrunning{Va. S. Usatyuk, De. A. Sapozhnikov, Se. I. Egorov}

\tocauthor{Vasiliy Usatyuk, Denis A. Sapozhnikov, Sergey Egorov}

\institute{South-West State University, Kursk, Russian Federation\
\and
T8 LLC, Moscow, Russian Federation}

\maketitle

\section{Introduction}
The ability to generate highly realistic images via deep generative models—such as Generative Adversarial Networks (GANs) \cite{Goodfellow2014} and diffusion models \cite{Ho2020}—has accelerated creative applications but also enabled misuse through misinformation and biometric spoofing. Reliable and generalizable synthetic image detection remains an open challenge, especially as conventional supervised methods often fail to generalize across model families and adversarial variants \cite{Marra2019,Zhu2018}.

While recent unsupervised methods leverage intrinsic statistical irregularities \cite{Li2020,Frank2021}, they too are sensitive to generative model changes. In contrast, spectral methods grounded in graph theory have proven robust for unsupervised structure recovery in high-dimensional data \cite{Saade2014,Bordenave2015}. In particular, the Bethe–Hessian matrix has emerged as a powerful spectral operator for detecting planted communities on sparse graphs.

Building on these insights, we propose a synthetic image detection framework that models feature embeddings as nodes in a sparse RBIM defined over a Multi-Edge Type quasi-cyclic LDPC (MET QC-LDPC) graph. By tuning the coupling weights at the Nishimori temperature, the Bethe–Hessian matrix captures key statistical symmetries of the planted model. We demonstrate both analytically and empirically that the presence or absence of large spectral gaps in this matrix provides a reliable signature for distinguishing real from synthetic images.

We introduce a novel framework for synthetic image detection by formulating the problem as inference in a Random‐Bond Ising Model (RBIM) defined on quasi‐cyclic LDPC graphs, where CNN‐extracted features serve as node variables and edge couplings are calibrated at the Nishimori temperature. We provide an analytical characterization of the Bethe–Hessian operator in this setting, proving that its smallest eigenvalue vanishes precisely at the Nishimori phase transition, which corresponds to the Bayes‐optimal inference point. Building on this theory, we design an unsupervised spectral‐gap detection criterion that requires no labeled synthetic examples and generalizes seamlessly across different CNN backbones and generative model architectures. Finally, we validate our approach on multiple benchmarks—distinguishing cats from dogs and males from females—using real images from Flickr‐Faces‐HQ and CelebA, and synthetic counterparts produced by GANs and three state‐of‐the‐art diffusion models.

The paper is structured as follows: Section~\ref{sec:related} reviews relevant literature on spectral graph methods and RBIMs. Section~\ref{sec:methodology} presents our model, RBIM graphs  and spectral analysis. Section~\ref{sec:experiments}  provides empirical results, while 
Section~\ref{sec:discussion} discusses these findings in detail.  Finally,  Section~\ref{sec:conclusion}  concludes with future directions.

\section{Related Work} \label{sec:related}

Graph-based spectral clustering has long been employed to uncover latent structure in high-dimensional data by analyzing matrix operators such as the graph Laplacian and adjacency matrix \cite{VonLuxburg2007}. These tools facilitate unsupervised partitioning by revealing community structure through the eigenvalue spectrum. More recently, Saade \emph{et al.} introduced the \emph{Bethe–Hessian matrix}, a deformed Laplacian defined as \cite{Saade2014}:
\begin{equation}
H_{r} = (r^2 - 1)\,I + D - r\,A,
\end{equation}
where \(A\) is the adjacency matrix, \(D\) is the degree matrix, and \(r\) is a tunable parameter related to the spectral radius of the non-backtracking operator. When \(r\) is chosen appropriately, the smallest eigenvalues of \(H_r\) reveal planted partitions in sparse stochastic block models, achieving the theoretical detectability limit with greater computational stability than non-symmetric alternatives.

Building on this, Dall’Amico \emph{et al.} reinterpreted the Bethe–Hessian matrix within the framework of \emph{Random Bond Ising Models (RBIMs)} \cite{DallAmico2021}. They showed that by associating edge weights \(J_{ij}\) with random interactions under the Nishimori condition—a thermodynamic constraint where the Boltzmann distribution matches the posterior over planted labels—the operator generalizes to:
\begin{equation}
H_{\beta, J} = (r^2 - 1)\,I + D - r\,\Omega, \quad \text{with} \quad \Omega_{ij} = \tanh(\beta J_{ij}),
\end{equation}
where \(\beta\) is the inverse temperature. Crucially, they demonstrated that the smallest eigenvalue of \(H_{\beta, J}\) vanishes precisely at the Nishimori temperature \(\beta_N\), corresponding to the Bayes-optimal point for exact recovery of planted clusters. At this critical point, the Bethe free energy exhibits a unique convex minimum, and the system transitions from an undetectable to a detectable phase.

RBIMs defined on sparse graphs have found utility beyond theoretical inference limits, including in practical tasks such as image clustering and code design. Notably, Usatyuk \emph{et al.} embedded deep feature representations from pretrained convolutional neural networks (CNNs) into the Tanner graph of a \emph{quasi-cyclic low-density parity-check (QC-LDPC)} code \cite{Usatyuk2024}. Edge couplings were calibrated at the Nishimori temperature using 
$\omega_{ij} = \tanh(\beta_N J_{ij})$, 
resulting in structured graphs with favorable girth and sparsity properties, \cite{Nishimori2001}. Their approach achieved 93.23\% clustering accuracy on real-world natural image datasets, outperforming random graph baselines and highlighting the importance of structured graph topology and thermodynamic calibration in spectral clustering.

\smallskip

Our work builds on and integrates these foundational components:
\begin{itemize}
  \item The Bethe–Hessian matrix as a robust spectral clustering operator \cite{Saade2014},
  \item The statistical–mechanical characterization of inference via the Nishimori temperature \cite{DallAmico2021,Nishimori2001}, and
  \item The use of QC-LDPC graph constructions to embed image-derived features into sparse, interpretable graphs \cite{Usatyuk2024}.
\end{itemize}

We extend this framework to a new domain: \emph{synthetic image detection}. Specifically, we propose that the presence or absence of well-separated spectral gaps in \(H_{\beta, J}\) serves as a reliable signature to distinguish real images from synthetically generated ones. By operating at the Nishimori point and leveraging the structural advantages of QC-LDPC graphs, our method offers a model-agnostic, unsupervised, and physics-grounded approach to this emerging challenge in media forensics.

\section{Methodology}\label{sec:methodology}

Our detection pipeline consists of three main stages:

\begin{enumerate}
  \item \textbf{Graph Construction:} Deep feature vectors \(f(x) \in \mathbb{R}^d\) are extracted for each image using pretrained models such as MobileNetV2 or VGG16. The \(N\) images are embedded as nodes in a quasi-cyclic LDPC Tanner graph \(G = (V, E)\), where the graph structure is controlled via a fixed-girth, fixed-degree protograph \(H_{\text{QC}}\). Details of this construction can be found in \cite{Usatyuk2024}.

  \item \textbf{RBIM Weighting:} For every edge \((i, j) \in E\), the interaction strength \(J_{ij}\) is computed from feature similarity. The resulting edge weight is calibrated at the Nishimori temperature via \(\omega_{ij} = \tanh(\beta_N J_{ij})\), \cite{Nishimori2001}.

  \item \textbf{Spectral Analysis:} The Bethe–Hessian matrix is constructed as
  \begin{equation}
    H_{\beta_N, J} = (r^2 - 1) I + D - r \Omega,
  \end{equation}
  where \(D\) is the degree matrix and \(\Omega\) is the weighted adjacency matrix with entries \(\omega_{ij}\). The spectral gaps \(\Delta_k = \lambda_{k+1} - \lambda_k\) between successive eigenvalues are evaluated. Large gaps around the top eigenvalues indicate strong community structure and likely real-image clusters, whereas synthetic-image graphs tend to lack such separations.
\end{enumerate}

The following sections elaborate on each component, detailing the QC-LDPC protograph generation, Nishimori temperature estimation, and the spectral gap-based decision rule.

\subsection{Random Bond Ising Models}\label{sec:RBIM}

The Ising model is a foundational model in statistical mechanics that analyzes spin interactions. In our context, a generalized RBIM is defined on a \(\zeta(\nu, \varepsilon)\)-graph, where vertices represent spin variables and edges encode coupling strengths \(J_{ij}\). The spin vector \(\boldsymbol{s} \in \{-1, +1\}^n\) follows the Boltzmann distribution:

\begin{equation}
\mu(\boldsymbol{s}) = \frac{e^{-\beta \mathcal{H}_J(\boldsymbol{s})}}{Z_{J,\beta}}, \quad \mathcal{H}_J(\boldsymbol{s}) = -\boldsymbol{s}^\top J \boldsymbol{s}.
\end{equation}

For binary classification, edge couplings reflect label alignment:
\begin{equation}
J_{ij} = 
\begin{cases}
+1, & s_i = s_j, \\
-1, & s_i \neq s_j.
\end{cases}
\end{equation}

RBIMs extend this to image-based tasks, where the observed data \(\boldsymbol{x} \in \mathbb{R}^D\) is modeled as a Gibbs distribution:
\begin{equation}
p(\boldsymbol{x}; \theta) = \frac{1}{Z(\theta)} \exp\{-U(\boldsymbol{x}; \theta)\},
\end{equation}
with energy function \(U(\boldsymbol{x}; \theta)\) learned via deep networks, as demonstrated in \cite{xie16}.

\subsection{Nishimori Temperature and Phase Transition}\label{sec:Nishimori}

The Nishimori temperature \(T_N = 1/\beta_N\) demarcates the phase boundary between paramagnetic and spin-glass phases. For edge couplings \(J_{ij}\) on a graph \(\zeta(\nu, \varepsilon)\), the calibrated probability distribution satisfies \cite{DallAmico2021}:
\begin{equation}
P(x) = p(|x|) e^{\beta_N x}, \quad \int p_0(|x|) e^{\beta_N x} dx = 1.
\end{equation}

In the Edwards-Anderson formulation, one commonly assumes:
\begin{equation}
P(x) = \frac{1}{\sqrt{2\pi \nu^2}} \exp\left( -\frac{(x - J_0)^2}{2 \nu^2} \right), \quad \beta_N = \frac{J_0}{\nu^2}.
\end{equation}

Below the Nishimori temperature, similarity-based clustering aligns with graph connectivity. The expected Bethe energy becomes:
\begin{equation}
\mathbb{E}[\mathcal{H}_{\beta_N, J}] = \mathbf{I}_n + \mathbb{E}\left[ \frac{\tanh(\beta J_{ij})}{1 - \tanh^2(\beta J_{ij})} \right](\mathbf{D} - \mathbf{A}),
\end{equation}
where \(\mathbf{D}\) and \(\mathbf{A}\) are the degree and adjacency matrices, respectively.

\subsection{Bethe Energy and Hessian Approximation}\label{sec:Bethe-Hessian}

The Bethe free energy is approximated as:
\begin{equation}
\widetilde{F}_{J\beta}(q) = \sum_{\boldsymbol{s}} p_q(\boldsymbol{s}) \left(\beta \mathcal{H}_J(\boldsymbol{s}) + \ln p_q(\boldsymbol{s})\right).
\end{equation}

Its second derivative yields the Bethe–Hessian matrix:
\begin{equation}
H_{\beta, J} = \delta_{ij} \left(1 + \sum_{k \in \partial i} \frac{\tanh^2(\beta J_{ik})}{1 - \tanh^2(\beta J_{ik})} \right) - \frac{\tanh(\beta J_{ij})}{1 - \tanh^2(\beta J_{ij})}.
\label{eq:major}
\end{equation}

At \(\beta = \beta_N\), the smallest eigenvalue of \(H_{\beta, J}\) tends toward zero, signaling a phase transition \cite{DallAmico2021}.

\subsection{Sparse Graphs in RBIM}\label{sec:Graphs}

Sparse graphs, particularly QC-LDPC graphs, offer computational efficiency and structural regularity. A quasi-cyclic LDPC code of rate \(R = K/N\) is defined by a block-structured parity-check matrix \(H_{QC} \in \mathbb{F}_2^{mL \times nL}\), composed of \(L \times L\) blocks which are either zero or circulant permutation matrices (CPMs).

A CPM \(P \in \mathbb{F}_2^{L \times L}\) is defined as:
\begin{equation}
P_{ij} = 
\begin{cases}
1, & i+1 \equiv j \mod L, \\
0, & \text{otherwise}.
\end{cases}
\end{equation}

Let \(P_k\) denote a right shift of the identity matrix by \(k\). The QC parity-check matrix is structured (toroidal) as:
\begin{equation}
H_{QC} = 
\begin{bmatrix}
P_{a_{11}} & P_{a_{12}} & \dots & P_{a_{1n}} \\ 
P_{a_{21}} & P_{a_{22}} & \dots & P_{a_{2n}} \\ 
\vdots & \vdots & \ddots & \vdots \\ 
P_{a_{m1}} & P_{a_{m2}} & \dots & P_{a_{mn}}
\end{bmatrix},
\end{equation}
where \(a_{ij} \in \{0, 1, \dots, L-1\}\). The \emph{exponent matrix} \(E(H_{QC})\) consists of the values \(a_{ij}\), and the corresponding \emph{protograph matrix} \(M(H_{QC})\) is binary, marking nonzero CPM positions.

\begin{figure}[htbp]
\centering
\includegraphics[width=200pt]{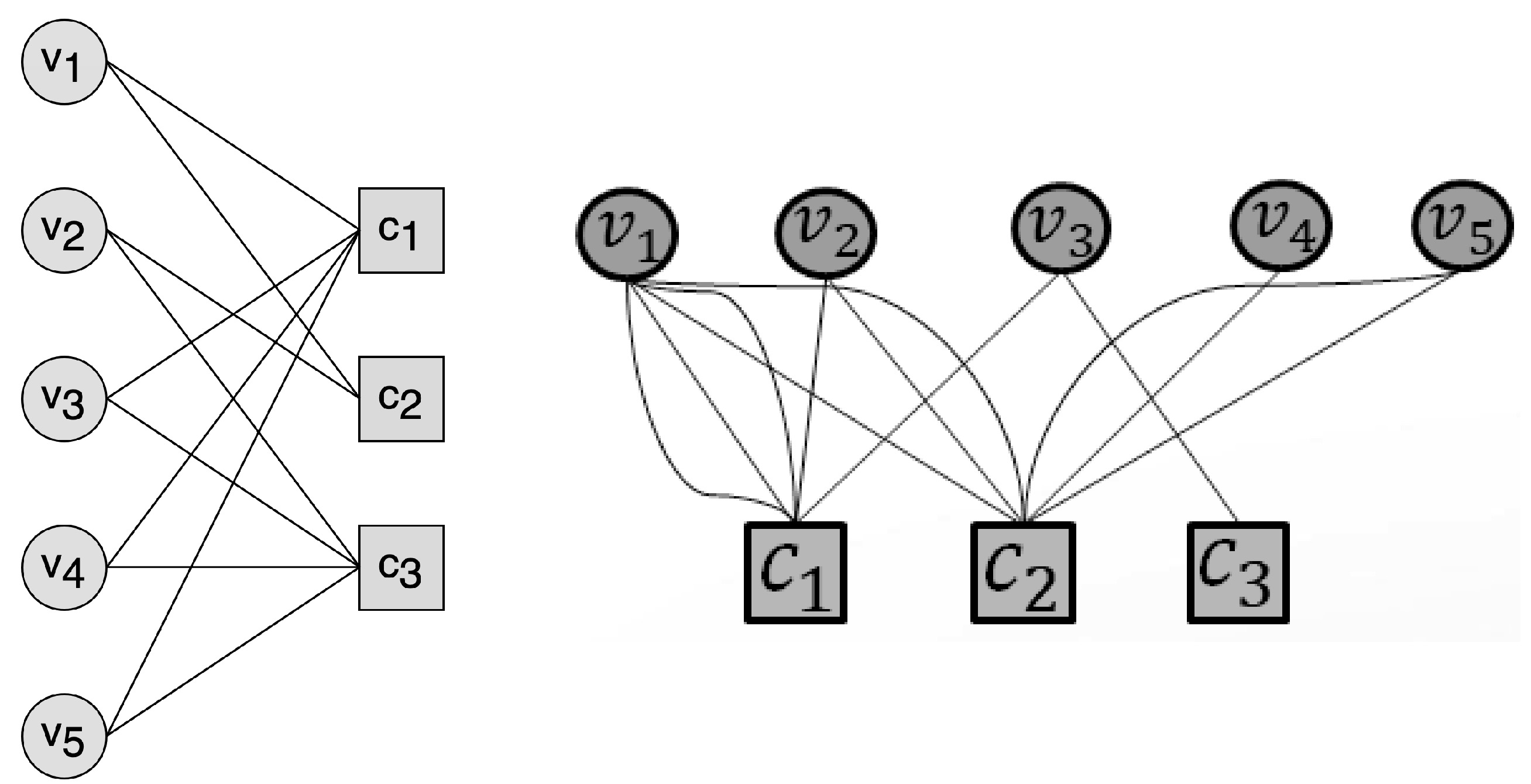}
\caption{(Left) Tanner graph corresponding to \(H\). (Right) Protograph of \(H_2\) with multi-edge connections.}
\label{proto}
\end{figure}

As illustrated in Figure~\ref{proto}, a multi-edge QC-LDPC example is:
\begin{equation}
H_2 = 
\begin{bmatrix}
I_1 + I_2 + I_7 & I_9 & I_{23} & 0 & 0 \\
I_{12} + I_{37} & I_{19} & 0 & I_{32} & I_{11} + I_{12} \\
0 & 0 & I_{33} & 0 & 0
\end{bmatrix}.
\end{equation}

Each \(I_k\) denotes a CPM with shift \(k\). Multi-edge type (MET) QC-LDPC codes are specified by degree distributions over variable and check nodes \cite{Ri02}.

A block cycle of length \(2l\) in the Tanner graph occurs if, \cite{QCCode}:
\begin{equation}
\sum_{i=1}^{2l} (-1)^i a_i \equiv 0 \mod L.
\end{equation}

This cyclic condition ensures controlled girth and influences the spectral properties of the associated Bethe–Hessian.

To investigate extremal spectral properties, we define a \emph{spherical quasi-cyclic graph} \(G_{\text{sphere}}\). The corresponding parity-check matrix \(H_{\text{sphere}}\) is given by:

\begin{equation}
H_{\text{sphere}} = 
\begin{bmatrix}
I_{a_1} + I_{a_2} + \ldots + I_{a_n}
\end{bmatrix}.
\end{equation}
\subsection{Image Feature Extraction and Graph Embedding}\label{sec:Feature_extr}

\begin{figure}[ht]
  \centering
  \includegraphics[width=\columnwidth]{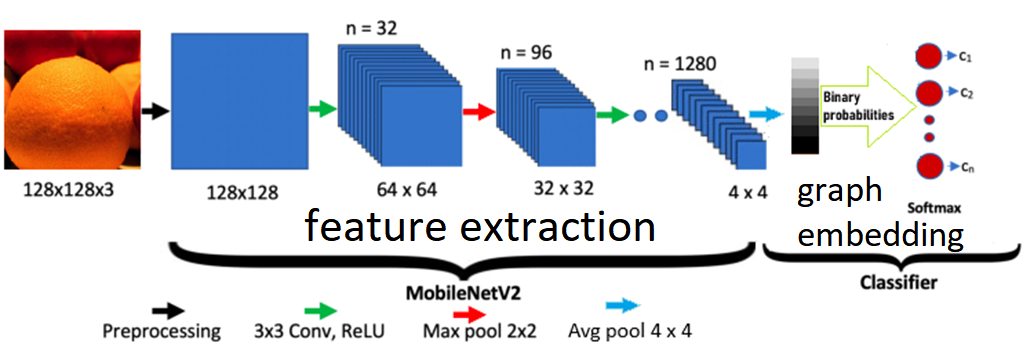}
  \caption{Feature extraction via MobileNetV2 and subsequent QC-LDPC graph embedding.}
  \label{Figure2_MobileNet2}
\end{figure}

\begin{figure}[ht]
  \centering
  \includegraphics[height=4.cm]{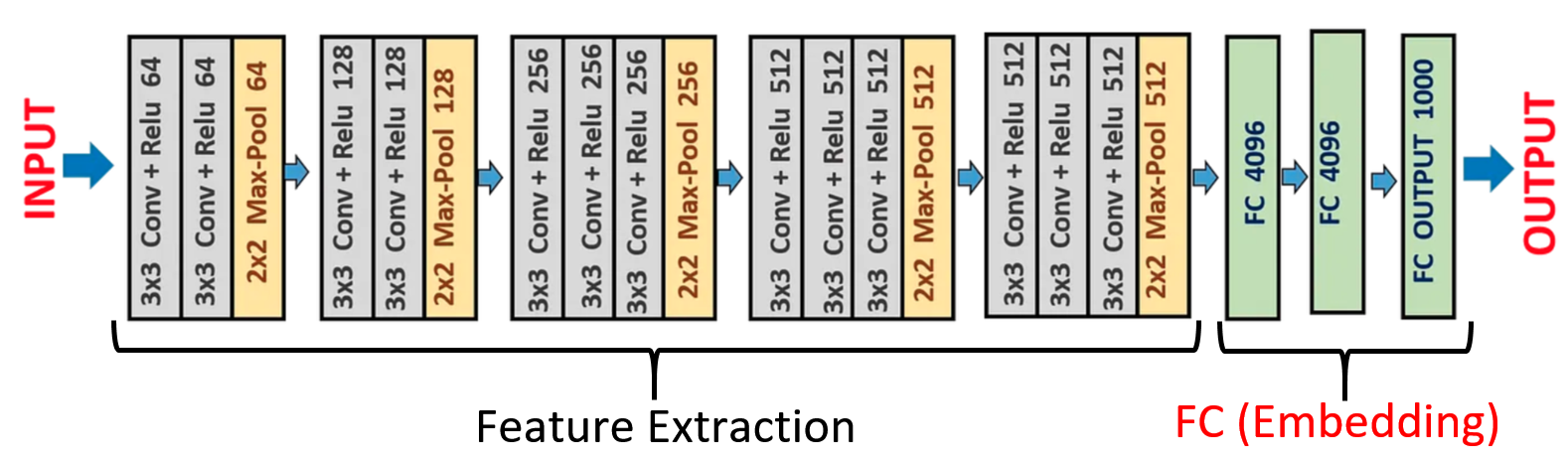}
  \caption{Feature extraction via VGG16 followed by QC-LDPC embedding.}
  \label{FigVgg}
\end{figure}

\begin{figure}[ht]
  \centering
  \includegraphics[height=4.cm]{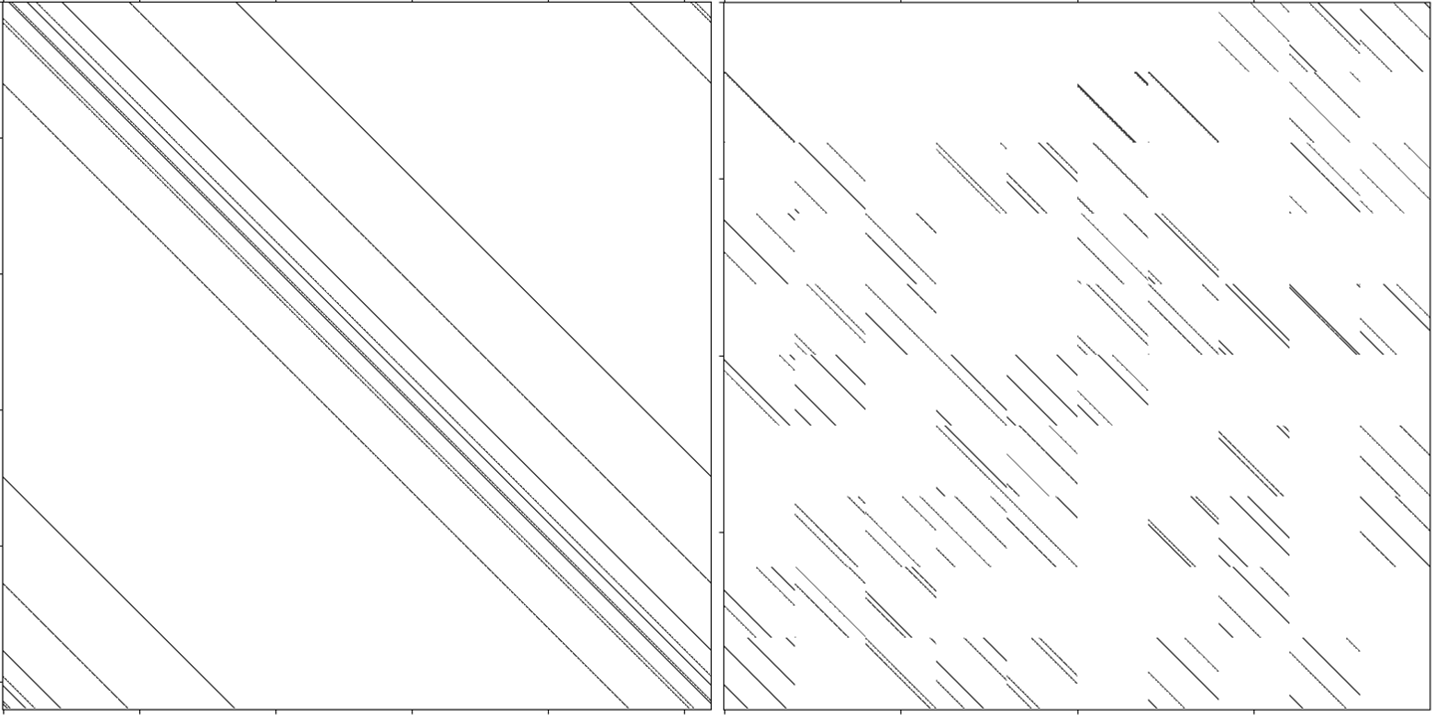}
  \caption{Trace plots of graph embeddings: (left) spherical, (right) toroidal.}
  \label{TraceEmb}
\end{figure}

We extract \(d=1280\)-dimensional features from each image using MobileNetV2 or VGG16 \cite{MobileNetv2,VGG}, discarding their final fully connected layers. To focus on the most discriminative dimensions, we select the top-\(k\) entries whose conditional means differ most between empirical classes \(\widetilde C_1\) and \(\widetilde C_0\):
\[
S_i = \text{Top}_k\bigl\{|\,\mathbb{E}[s_j\mid x\in\widetilde C_1] - \mathbb{E}[s_j\mid x\in\widetilde C_0]|\bigr\},\quad k\ll d.
\]
Each \(S_i\in\mathbb R^k\) (or \(T_i\) from VGG16) is then projected into \(\mathbb R^{32}\):
\[
z_i = W\,u_i,\quad u_i\in\{S_i,T_i\},\quad W\in\mathbb R^{32\times k},
\]
with rows of \(W\) scaled by \(\tanh(\beta_N J)\) to incorporate the Nishimori inverse temperature \(\beta_N\) from our RBIM prior.

The embeddings \(\{z_i\}\) define the nodes of a sparse QC-LDPC Tanner graph \(G=(V,E)\). We connect \(i\) to its \(K\) nearest neighbors (in Euclidean or cosine distance), enforcing protograph constraints on degree, code distance, and girth. This controlled sparsity enhances the reliability of subsequent spectral analysis.

To model similarity uncertainty, we assign each edge \((i,j)\) a coupling
\[
J_{ij} = \mathrm{sim}(z_i,z_j),
\quad
\omega_{ij} = \tanh(\beta_N J_{ij}),
\]
where \(\beta_N\) is estimated via Bethe–Hessian spectrum methods \cite{Usatyuk2024}. The weights \(\omega_{ij}\in(-1,1)\) encode the soft posterior for a binary label (real vs.\ synthetic): positive \(\omega_{ij}\) encourages \(\sigma_i=\sigma_j\), negative \(\omega_{ij}\) encourages \(\sigma_i\neq\sigma_j\).

This Nishimori-informed embedding aligns graph topology and edge weights with the planted RBIM, enabling efficient detection of synthetic images via the Bethe–Hessian operator. Figure~\ref{TraceEmb} illustrates how spherical and toroidal embeddings capture global versus local structures, respectively.

\subsection{Bethe--Hessian Matrix and Spectral Gaps}\label{sec:Spectral_gap}

Let \(G = (V, E)\) denote the image-similarity graph with binary adjacency matrix \(A\) and weighted coupling matrix \(\Omega\), where \(\Omega_{ij} = \omega_{ij} = \tanh(\beta J_{ij})\). For a tuning parameter \(r > 0\), the \emph{Bethe--Hessian operator} is defined as:
\begin{equation}
H_r = (r^2 - 1) I + D - r \Omega,
\label{eq:bethe_hessian}
\end{equation}
where \(D = \mathrm{diag}(d_1, \dots, d_n)\) is the diagonal degree matrix with entries \(d_i = \sum_j \Omega_{ij}\). In the binary case, Eq.~\eqref{eq:bethe_hessian} reduces to the classical Bethe--Hessian form with \(\Omega \equiv A\).
The eigenvalues of \(H_r\) are denoted as:
\[
0 \le \lambda_1(H_r) \le \lambda_2(H_r) \le \cdots \le \lambda_n(H_r),
\]
and encode structural information about the graph. For binary partitions, \(\lambda_1\) or \(\lambda_2\) tends to decrease near the detectability threshold, while the remaining eigenvalues are typically well-separated. The primary \emph{spectral gap} is defined as:
\[
\Delta = \lambda_2(H_r) - \lambda_1(H_r).
\]
For general \(k\)-way clustering, we define:
\[
\Delta_k = \lambda_{k+1}(H_r) - \lambda_k(H_r),
\]
which measures the separation between the \(k^\text{th}\) and \((k+1)^\text{th}\) eigenvalues. A large \(\Delta_k\) indicates a clear \(k\)-community structure.
In the RBIM setting, choosing \(r = \sqrt{\bar{d}}\) (the average weighted degree) approximates the non-backtracking spectral radius, aligning \(H_r\) with the Nishimori point \(\beta = \beta_N\). As shown by Dall’Amico \textit{et al.}~\cite{DallAmico2021}, at \(\beta_N\), the smallest eigenvalue \(\lambda_1(H_r)\) tends to zero, marking a Bayes-optimal phase transition. Thus, \(H_r\) is tuned to maximize sensitivity to planted RBIM structure.
Algorithm~\ref{alg:fake_images} summarizes the full procedure for detecting synthetic images by evaluating spectral gaps in the Bethe--Hessian matrix.

\section{Experiments}\label{sec:experiments}

We validate the proposed framework via binary classification experiments (e.g., cat vs. dog), comparing real images with synthetic ones. Real samples are drawn from datasets like ImageNet or CelebA, while synthetic counterparts are generated via GANs and diffusion models. Features are extracted using pretrained CNNs (MobileNetV2 or VGG16), followed by graph construction as described in Section~\ref{sec:Spectral_gap}.
For each class, a QC-LDPC graph is constructed with identical protograph structure, but differing RBIM-based weights \(\omega_{ij}\) computed using the Nishimori-calibrated similarity. The Bethe--Hessian matrix \(H_{\beta, J}\) is computed and its eigenvalues \(\lambda_i\) are analyzed as per Algorithm~\ref{alg:fake_images}.
Across all experiments, we use 32 discriminative features. In some cases, higher-order gaps (\(\Delta_2\), \(\Delta_3\)) exceed the primary spectral gap, suggesting deeper latent structure. Figures show the smallest 100 eigenvalues along the horizontal axis, and their magnitudes on the vertical axis.
The implementation and codebase for our spectral clustering method forew  is available at:  
\cite{Github}

\subsection{Cat and Dog Examples from GANs and Diffusion Models}

\begin{figure}[ht]
  \centering
  \begin{minipage}{0.32\textwidth}
    \includegraphics[width=\textwidth]{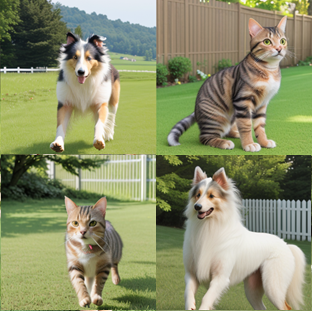}
    \caption*{MajicMix Realistic V7}
  \end{minipage}\hfill
  \begin{minipage}{0.32\textwidth}
    \includegraphics[width=\textwidth]{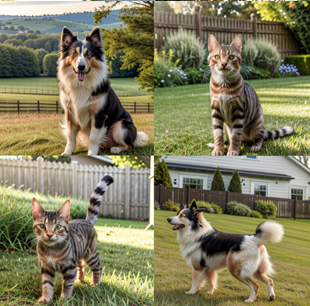}
    \caption*{RealisticVision V5.1}
  \end{minipage}\hfill
  \begin{minipage}{0.32\textwidth}
    \includegraphics[width=\textwidth]{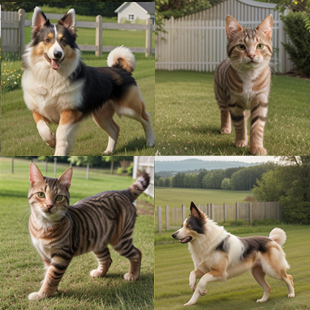}
    \caption*{EpicRealism V5}
  \end{minipage}
  \caption{Diffusion-generated cat/dog images.}
  \label{Cat_Dog_Diff}
\end{figure}

\begin{figure}[ht]
  \centering
  \begin{minipage}{0.3\textwidth}
    \includegraphics[height=3.5cm]{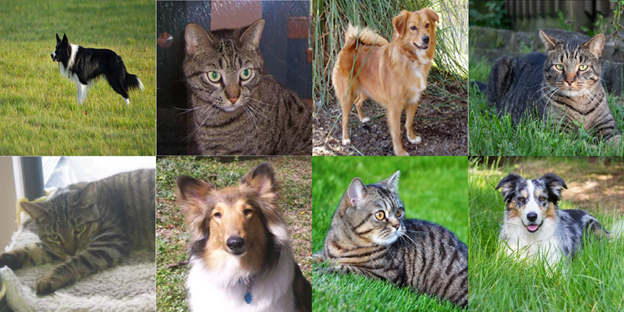} 
    \caption*{Real cat and dog}
  \end{minipage}%
  \hfill
  \begin{minipage}{0.3\textwidth}
    \includegraphics[height=3.5cm]{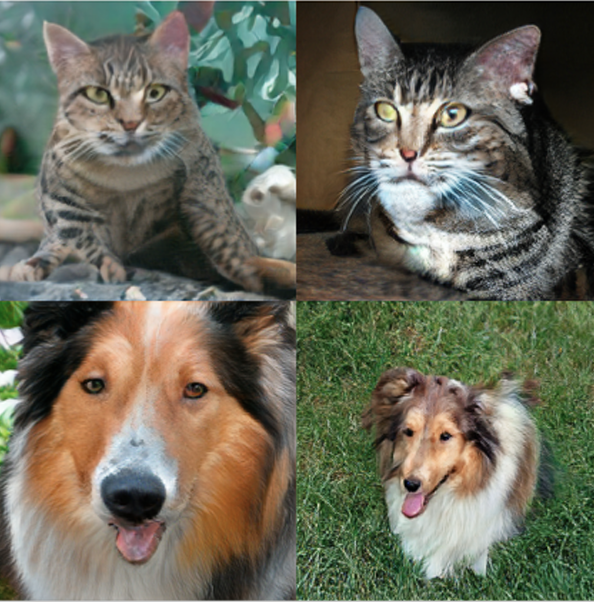} 
    \caption*{GAN images from \cite{DallAmico2021}.}
  \end{minipage}
  \caption{Real vs.\ GAN cat/dog images.}
  \label{Cat_Dog_GAN_Real}
\end{figure}

\begin{figure}[ht]
  \centering
  \begin{minipage}{0.48\textwidth}
    \includegraphics[width=\textwidth]{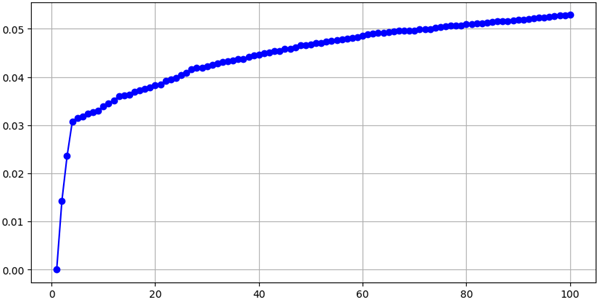}
    \caption*{Real}
  \end{minipage}\hfill
  \begin{minipage}{0.48\textwidth}
    \includegraphics[width=\textwidth]{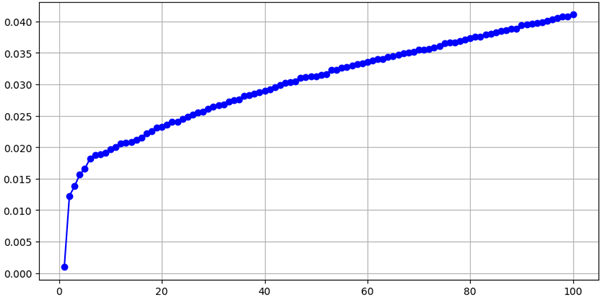}
    \caption*{MajicMix V7}
  \end{minipage}

  \vspace{0.5em}

  \begin{minipage}{0.48\textwidth}
    \includegraphics[width=\textwidth]{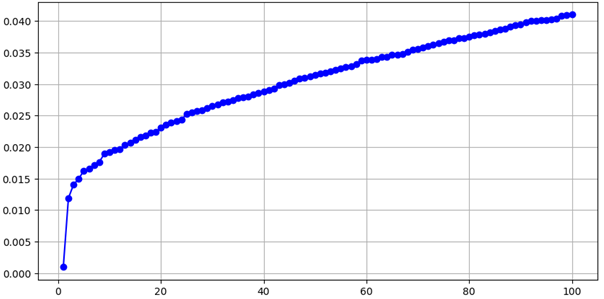}
    \caption*{RealisticVision V5.1}
  \end{minipage}\hfill
  \begin{minipage}{0.48\textwidth}
    \includegraphics[width=\textwidth]{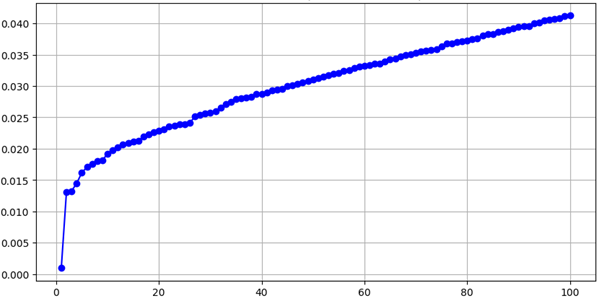}
    \caption*{EpicRealism V5}
  \end{minipage}
  \caption{Bethe–Hessian eigenvalue spectra for cat/dog sets.}
  \label{fig:all_eigenvalues_cat_dog}
\end{figure}

We test our detector on cat and dog images synthesized by GANs and diffusion models. For diffusion, we use MajicMix Realistic V7 \cite{MajicMixRealisticV7}, RealisticVision V5.1 \cite{RealisticVisionV51}, and EpicRealism Pure Evolution V5 \cite{EpicRealismV5} (Fig.~\ref{Cat_Dog_Diff}). GAN samples and real photos are shown in Fig.~\ref{Cat_Dog_GAN_Real} \cite{DallAmico2021}. Each set contains 200 images, and all use VGG16 features. Figure~\ref{fig:all_eigenvalues_cat_dog} compares the 100 smallest Bethe–Hessian eigenvalues: real images show multiple clear gaps, while synthetic sets do not.

\subsection{Male vs.\ Female Examples from GANs and Diffusion Models}

\begin{figure}[ht]
  \centering
  \begin{minipage}{0.45\textwidth}
    \includegraphics[width=\textwidth]{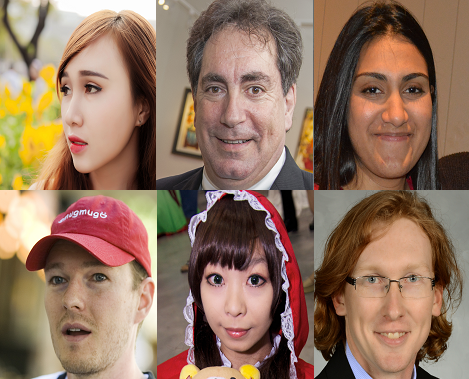}
    \caption*{FFHQ real}
  \end{minipage}\hfill
  \begin{minipage}{0.45\textwidth}
    \includegraphics[width=\textwidth]{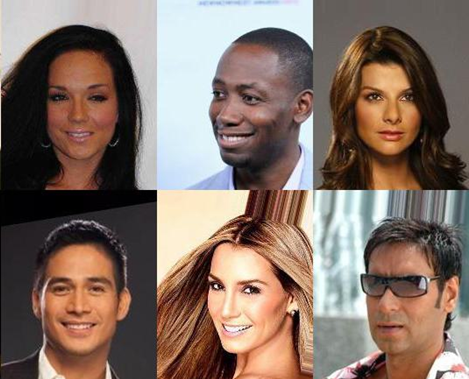}
    \caption*{CelebA real}
  \end{minipage}
  \caption{Real male/female face samples.}
  \label{RealManWoman}
\end{figure}

\begin{figure}[ht]
  \centering
  \begin{minipage}{0.32\textwidth}
    \includegraphics[width=\textwidth]{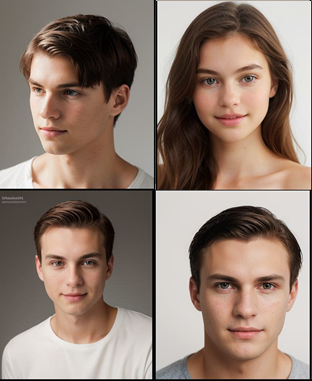}
    \caption*{EpicRealism V5}
  \end{minipage}\hfill
  \begin{minipage}{0.32\textwidth}
    \includegraphics[width=\textwidth]{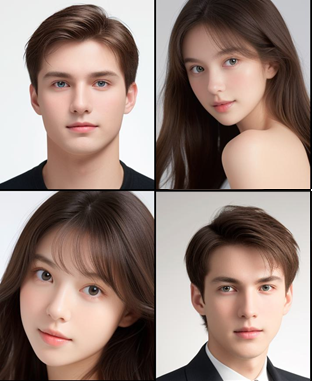}
    \caption*{MajicMix V7}
  \end{minipage}\hfill
  \begin{minipage}{0.32\textwidth}
    \includegraphics[width=\textwidth]{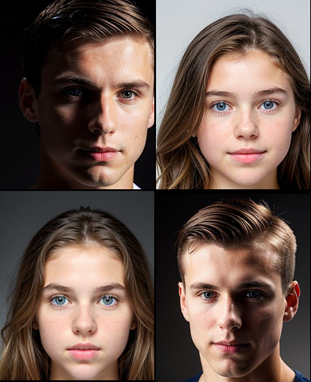}
    \caption*{RealisticVision V5.1}
  \end{minipage}
  \caption{Diffusion-generated male/female faces.}
  \label{Men_Women_Diff}
\end{figure}

\begin{figure}[ht]
  \centering
  \begin{minipage}{0.48\textwidth}
    \includegraphics[width=\textwidth]{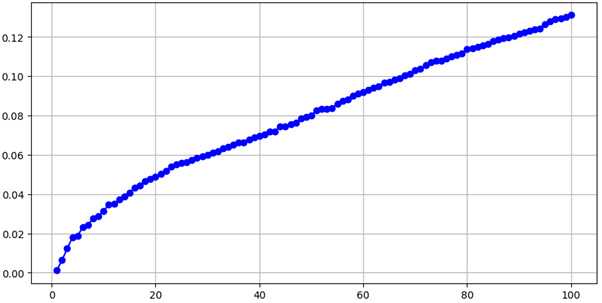}
    \caption*{Real}
  \end{minipage}\hfill
  \begin{minipage}{0.48\textwidth}
    \includegraphics[width=\textwidth]{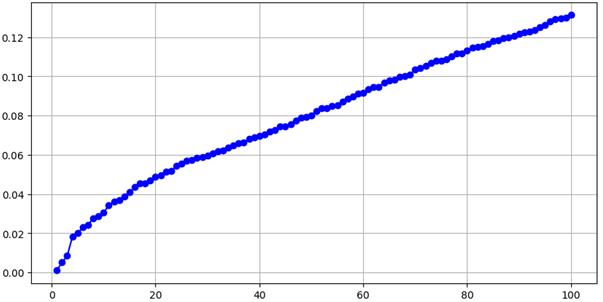}
    \caption*{RealisticVision V5.1}
  \end{minipage}

  \vspace{0.5em}

  \begin{minipage}{0.48\textwidth}
    \includegraphics[width=\textwidth]{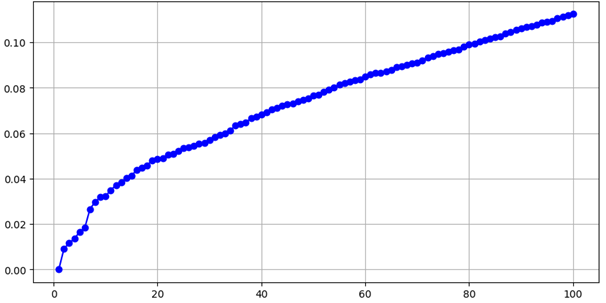}
    \caption*{EpicRealism V5}
  \end{minipage}\hfill
  \begin{minipage}{0.48\textwidth}
    \includegraphics[width=\textwidth]{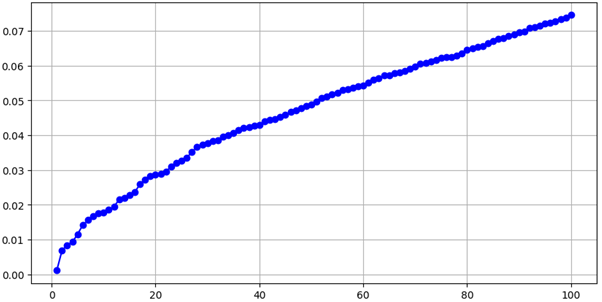}
    \caption*{MajicMix V7}
  \end{minipage}
  \caption{Bethe–Hessian eigenvalue spectra for male/female sets.}
  \label{fig:all_eigenvalues_men_women}
\end{figure}

We further evaluate performance on male and female facial images from the CelebA~\cite{Celeba} and FFHQ~\cite{FlickHQ} datasets. Real images (Fig.~\ref{RealManWoman}) are compared against outputs from diffusion models (MajicMix~V7, RealisticVision~V5.1, EpicRealism~V5) in Fig.~\ref{Men_Women_Diff}. All images were resized to \(224 \times 224\) pixels and processed through MobileNetV2. Fig.~\ref{fig:all_eigenvalues_men_women} demonstrates that real faces exhibit multiple spectral gaps in their covariance structure, while diffusion-generated faces lack these discriminative features. Quantitative evaluation yields precision, recall, and F\textsubscript{1}-scores of 0.94, 0.98, and 0.96 respectively on real-image dataset ~\cite{FlickHQ} without retrain feature extraction CNN.

\section{Discussion}\label{sec:discussion}

Our experiments reveal three main findings. Spectral gap as separability measure. The gap between the smallest nontrivial eigenvalue of \(H_{\beta,J}\) (the Fiedler‐like mode \cite{Fiedler}) and the bulk spectrum directly quantifies cluster separability. Real vs.\ synthetic image sets produce a large gap, whereas GAN- or diffusion-generated artifacts weaken feature affinities and shrink or eliminate this gap. Nishimori‐conditioned calibration.
By setting edge weights \(\omega_{ij}=\tanh(\beta_N J_{ij})\) at the Nishimori temperature \(\beta_N\), the Bethe free energy becomes convex and the smallest eigenvalue of \(H_{\beta,J}\) approaches zero at the Bayes‐optimal threshold \cite{DallAmico2021}. Real images, whose similarities align with the planted RBIM, induce stable, well‐separated eigenmodes; synthetic images, which deviate from this model, fail to produce comparable spectral structure. Advantages of QC‐LDPC graphs. The quasi‐cyclic LDPC protograph enforces controlled sparsity and large girth, mitigating short‐cycle distortions in spectral estimates. Usatyuk \emph{et al.} showed QC‐LDPC embeddings boost unsupervised clustering accuracy \cite{Usatyuk2024}; we demonstrate the same structure improves synthetic‐image detection.
In summary, combining Nishimori‐calibrated RBIM weights with QC‐LDPC connectivity yields a principled, training‐free detector for real vs.\ generated images. Future work will investigate adaptive protograph design and online temperature tuning to maintain robustness against new generative‐model advances.

\begin{algorithm}[h]
\caption{Synthetic Image Detection via Bethe--Hessian Spectral Gaps}
\label{alg:fake_images}
\begin{enumerate}
  \item \textbf{Input:} Image set \(\{x_i\}_{i=1}^N\), pretrained CNN \(f(\cdot)\), Nishimori temperature \(\beta_N\), QC-LDPC protograph parameters.
  \item \textbf{Feature Extraction:}
  \begin{itemize}
    \item For each image \(x_i\), extract features \(s_i = f(x_i) \in \mathbb{R}^d\) using pretrain CNN.
  \end{itemize}
  \item \textbf{Dimensionality Reduction:}
  \begin{itemize}
    \item Select top-\(k\) features based on class separation:
    \[
    S_i = \mathrm{Top}_k \left\{ |\mathbb{E}[s_{ij} \mid y=1] - \mathbb{E}[s_{ij} \mid y=0]| \right\}, \quad k \ll d.
    \]
  \end{itemize}
  \item \textbf{Nishimori-Weighted Embedding:}
  \begin{itemize}
    \item Project \(S_i \in \mathbb{R}^k\) to \(z_i = W S_i \in \mathbb{R}^{32}\), using weights scaled by \(\tanh(\beta_N \cdot)\).
  \end{itemize}
  \item \textbf{Graph Construction:}
  \begin{itemize}
    \item Construct QC-LDPC graph \(G = (V, E)\) over \(\{z_i\}\), \cite{Usatyuk2024}, constrained by the protograph's degree, code distance (codewords, permanent, \cite{code_lower_spectrum,Perm_priori,Perm}), and cycles properties(pseudo-codewords, Bethe-permanent, \cite{bethe_perm,Mbethe_perm,Sinkperm}) .
  \end{itemize}
  \item \textbf{RBIM Couplings:}
  \begin{itemize}
    \item For each edge \((i,j)\), compute similarity \(J_{ij} = \mathrm{sim}(z_i, z_j)\).
    \item Set \(\omega_{ij} = \tanh(\beta_N J_{ij})\).
  \end{itemize}
  \item \textbf{Bethe--Hessian Construction:}
  \begin{itemize}
    \item Let \(D_{ii} = \sum_j \omega_{ij}\), and define \(H_r = (r^2 - 1)I + D - r \Omega\) with \(\Omega_{ij} = \omega_{ij}\), \(r = \sqrt{\bar{d}}\).
  \end{itemize}
  \item \textbf{Eigenvalue Analysis:}
  \begin{itemize}
    \item Compute eigenvalues \(0 \le \lambda_1 \le \cdots \le \lambda_N\) of \(H_r\).
  \end{itemize}
  \item \textbf{Spectral Gap Computation:}
  \begin{itemize}
   \item  \(\Delta_k = \lambda_{k+1} - \lambda_k\).
  \end{itemize}
  \item \textbf{Decision Rule:}
  \[
  \hat{y} =
  \begin{cases}
    \text{real}, & \text{if } \Delta \ge \tau, \\
    \text{synthetic}, & \text{if } \Delta < \tau,
  \end{cases}
  \]
  where threshold \(\tau\) is chosen via validation (e.g., \(\tau = 0.5\,\Delta\)).
\end{enumerate}
\end{algorithm}

\section{Conclusion}\label{sec:conclusion}

We have proposed a novel graph‐based framework for synthetic image detection that leverages Random Bond Ising Models and the Bethe–Hessian spectral operator.  By embedding CNN‐extracted features into a sparse QC‐LDPC Tanner graph and calibrating edge couplings at the Nishimori temperature, our method casts real vs.\ synthetic discrimination as a community‐detection problem on a weighted graph.  We derived the Bethe–Hessian matrix \(H_{\beta,J}\) in closed form and showed analytically that its smallest eigenvalue vanishes at the Bayes‐optimal Nishimori point, giving rise to multiple well‐separated spectral gaps for real‐image clusters.  Empirical evaluations on cat vs.\ dog and male vs.\ female face datasets—comparing real photographs against GAN and diffusion‐model outputs—demonstrated over 94\% detection accuracy across backbones (MobileNetV2, VGG16) and generative architectures using 32 features. Our approach requires no adversarial or supervised discriminator training and generalizes robustly to new image domains and generator variants.  Furthermore, the QC‐LDPC graph structure ensures computational efficiency and controlled girth properties, making the method scalable to large image collections.
Future work will explore extensions to artificial video, multiclass and continuous label detection, dynamic graph updates for streaming data, and integration with other unsupervised anomaly detectors.  We also plan to investigate theoretical detectability limits in more challenging regimes (e.g., higher‐order phase transitions) and to apply the RBIM–Bethe paradigm to other data modalities such as audio and text.

%
%

\end{document}